\documentclass{bmvc2k}
 \usepackage{amsmath}
 \usepackage{amsfonts}
\usepackage{amssymb}
\usepackage{pifont}% http://ctan.org/pkg/pifont
\newcommand{\cmark}{\ding{51}}%
\newcommand{\xmark}{\ding{55}}%

\title{Context-Aware Trajectory Prediction in Crowded Spaces}

\addauthor{Federico Bartoli}{federico.bartoli@unifi.it}{1}
\addauthor{Giuseppe Lisanti}{giuseppe.lisanti@unifi.it}{1}
\addauthor{Lamberto Ballan\textsuperscript{1,}}{lamberto.ballan@unipd.it}{2}
\addauthor{Alberto Del Bimbo}{alberto.delbimbo@unifi.it}{1}

\addinstitution{
Media Integration and\\Communication Center\\
University of Florence, Italy\vspace{4pt}
}
\addinstitution{
Department of Mathematics\\``Tullio Levi-Civita''\\
University of Padova, Italy
}

\runninghead{Bartoli \MakeLowercase{et al.}}{Context-Aware Trajectory Prediction}

\def\etal{\emph{et al}\bmvaOneDot}

\graphicspath{{./img/}}
 
\begin{document}

\maketitle

\begin{abstract}
Human motion and behaviour in crowded spaces is influenced by several factors, such as the dynamics of other moving agents in the scene, as well as the static elements that might be perceived as points of attraction or obstacles. In this work, we present a new model for human trajectory prediction which is able to take advantage of both human-human and human-space interactions. The future trajectory of humans, are generated by observing their past positions and interactions with the surroundings. To this end, we propose a ``context-aware'' recurrent neural network LSTM model, which can learn and predict human motion in crowded spaces such as a sidewalk, a museum or a shopping mall.
We evaluate our model on a public pedestrian datasets, and we contribute a new challenging dataset that collects videos of humans that navigate in a (real) crowded space such as a big museum.
Results show that our approach can predict human trajectories better when compared to previous state-of-the-art forecasting models.
\end{abstract}

%-------------------------------------------------------------------------
\section{Introduction}
\label{sec:intro}
People usually move in a crowded space, having a goal in mind, such as moving towards another subject active in the scene or reaching a particular destination. These might be, for example, a shopping showcase, a public building or a specific artwork in a museum. While doing so, they are able to consider several factors and adjust their path accordingly.
A person can adapt her/his path depending on the \emph{space} in which is moving, and on the other \emph{humans} that are walking around him. In the same way, an obstacle can slightly modify the trajectory of a pedestrian, while some other elements may constrain his path. % (e.g. sidewalk).

Although human behaviour understanding and people tracking have a long tradition in computer vision literature \cite{Pellegrini2009,kuettel2010cvpr,zen2011cvpr,taixe2014cvpr,solera2015iccv,Xie2013}, in the recent years we observe increasing interest also in predictive models \cite{Kitani2012,Walker2014,Gavrila2014,alahi2016lstm,ballan2016eccv,Walker2016}.
Observing how persons navigate a crowded scenario, and being able to \emph{predict} their future steps, is an extremely challenging task that can have key applications in robotic, smart spaces and automotive~\cite{luber2010people,trautman2013robot,alahi2014cvpr,ashesh2015iccv,forces2016eccv}.
%As an example, such a model could be used to observe a sidewalk and predict whether a person is going to suddenly cross the street and avoid collisions.

Predicting the motion of human targets, such as pedestrians, or generic agents, such as cars or robots, is a very challenging and open problem. %has recently attracted increasing interest from the computer vision community. 
Most of the existing work in this area address the task of predicting the trajectory of a target, by inferring some properties of the scene or trying to encode the interactive dynamics among observed agents.
Trajectory prediction is then achieved by modelling and learning \emph{human-space} \cite{Kitani2012,ballan2016eccv,huang2016tip} or \emph{human-human} interactions \cite{suav2016eccv,alahi2016lstm,yoo2016,kitani2017gt,vemula2017robot}.
Pioneering works have tried to parameterise human behaviours with hand-crafted models such as social forces \cite{helbing1995social,Pellegrini2009,Yama2011}, while more recent approaches have attempted to infer these behaviours in a data-driven fashion \cite{alahi2016lstm}.
This idea has proven to be extremely successful for improving performance in multi-target tracking applications, and short-term trajectory prediction. In fact, being able to take into account the interactions of nearby agents, is extremely important to avoid collisions in a crowded scenario \cite{Pellegrini2009,suav2016eccv,alahi2016lstm}.
At the same time, a prior knowledge about the interactions between a specific target and the static elements of the scene (e.g. location of sidewalks, buildings, trees, etc.), is essential to obtain reliable prediction models \cite{Kitani2012,ballan2016eccv}.
However, a main limitation of these models is that all of them attempt only to model human-human or human-space interactions.

\begin{figure*}
\centering
\includegraphics[width=.85\columnwidth]{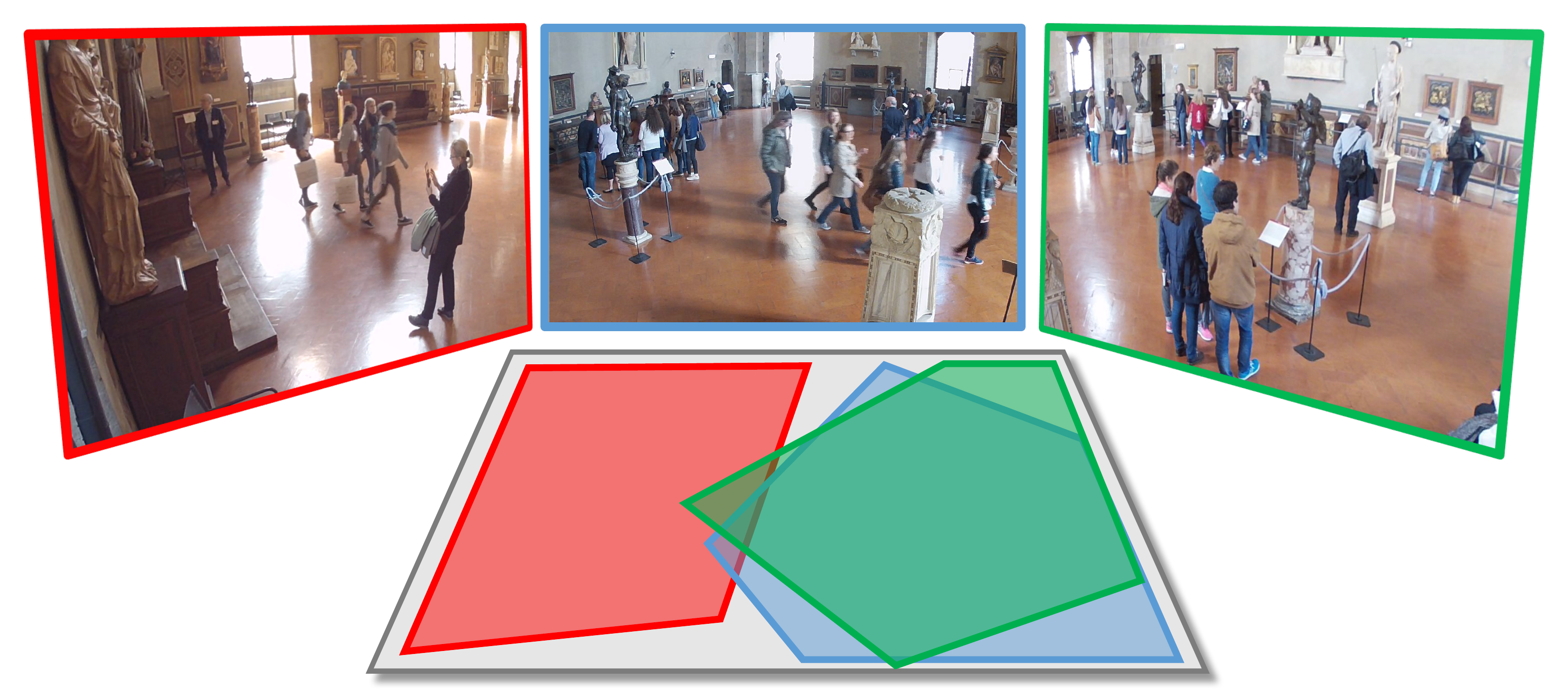}
\caption{We aim to learn and predict human behaviours in crowded spaces. To this end, we have collected a new dataset from multiple cameras in a big museum. In contrast to previous ones, our dataset allows experiments in a real crowded scenario where people interact not only with each others, but also with the space which is characterized by a rich semantic.}
\label{fig:pullfigure}
\end{figure*}

In this paper, we introduce a novel approach for human trajectory prediction which is based on a ``context-aware'' recurrent neural network model. 
%Our model is able to take into account both human-human and human-space interactions.
Each person trajectory is modelled through a Long-Short Term Memory (LSTM) network. In order to take into account for interactions with other humans and/or with the space, we extend the Social-LSTM model that has been recently proposed by Alahi \etal~\cite{alahi2016lstm}, by defining a ``context-aware'' pooling that allows our model to consider also the static objects in the neighborhood of a person.
The proposed model observes the past positions of a human and his interactions with the surroundings in order to predict his near future trajectory. Results demonstrate that considering both human-human and human-space interactions is fundamental for obtaining an accurate prediction of a person' trajectory.

The main contributions of this paper are two-fold: \emph{i}) we introduce a trajectory prediction model based on an LSTM architecture and a novel ``context-aware'' pooling which is able to learn and encode both human-human and human-space interactions; \emph{ii}) we demonstrate the effectiveness of the proposed model in comparison to previous state of the art approaches, such as the recent Social-LSTM \cite{alahi2016lstm}, on the UCY dataset \cite{crowdbyex} and a new challenging dataset, called MuseumVisits, that we will release publicly.

%-------------------------------------------------------------------------
%\section{Related Work}\label{sec:related}
%\begin{itemize}
%item{Activity recognition and trajectory analysis: \cite{Pellegrini2009,kuettel2010cvpr,zen2011cvpr,taixe2014cvpr,solera2015iccv,Yama2011,Xie2013}}
%\item{Predicting motions and trajectory forecasting: \cite{Kitani2012,Gavrila2014,Walker2014,alahi2014cvpr,alahi2016lstm,suav2016eccv,ballan2016eccv,kitani2017gt}}
%\end{itemize}

%-------------------------------------------------------------------------
\section{Approach}\label{sec:approach}
%\textcolor{red}{Predicting as accurately as possible the trajectory of a person seems an impossible task, mainly because of the fact that no one knows a priori where a person is going. However, several factors in the observed scene, both humans and non, may help in this task. For this reason we believe that modeling also the context in which a person is moving is fundamental.}

In this section, we describe our ``context-aware'' model for trajectory prediction. Our key assumption is that the behaviour of each person is strongly influenced by his interactions with the context, both in terms of static elements of the scene and dynamics agents active in the same scenario (such as other pedestrians). To this end, we model each person's trajectory with an LSTM network. Our work builds on top of the recent Social-LSTM model \cite{alahi2016lstm}, but we introduce a more general formulation which can include both human-human and human-space interactions.

\subsection{Context-Aware LSTM}
Given a video sequence, each trajectory is expressed in the form of spatial coordinates, such that  $X^i_t=(x^i_t,y^i_t)$ represent the location of the $i$-th person at time instant $t$. 
We use an LSTM network to represent the i-th person trajectory, as follows:
\begin{align}
\begin{pmatrix}
  \mathbf{i}^i_t\\
  \mathbf{f}^i_t\\
  \mathbf{o}^i_t\\
  \mathbf{\tilde{c}}^i_t
 \end{pmatrix}&=
 \begin{pmatrix}
  \sigma\\
  \sigma\\
  \sigma\\
  \tanh
 \end{pmatrix}
\begin{bmatrix}
 W
  \begin{pmatrix}
   \mathbf{h}^i_{t-1}\\
   \mathbf{x}^i_t\\
 \end{pmatrix} +
 b^i
 \end{bmatrix},\\ 
\mathbf{x}^i_t&=\phi(X^i_t,W_x),\label{eq:xt} \\ 
\mathbf{c}^i_t&=\mathbf{f}^i_t\odot \mathbf{c}^i_{t-1}+\mathbf{i}^i_{t}\odot \mathbf{\tilde{c}}^i_t,\\
\mathbf{h}^i_t&=\mathbf{o}^i_t \odot \tanh(\mathbf{c}^i_t),
\end{align}
where $\mathbf{x}^i_t\in \mathbb{R}^n$ represents the input data, $\mathbf{h}_{t}^i\in \mathbb{R}^D$ is the output state and $\mathbf{c}_t \in \mathbb{R}^D$ is the hidden  state of the LSTM at time $t$. The input $\mathbf{x}_t^i$ is obtained by applying the ReLU function $\phi(\cdot)$ to the spatial coordinates $X^i_t$ and the weight matrix $W_x\in \mathbb{R}^{N\times 2}$. 

At training time\footnote{As in previous solutions~\cite{alahi2016lstm,Pellegrini2009}, we assume that a set of trajectories for all the persons observed in a scene is available a priori for training.}  the current input and the output from the previous instant are updated according to the weights $W\in \mathbb{R}^{(4D) \times (n+D)}$  and the bias term $b\in \mathbb{R}^{(n+D)}$.
Then, the updated vectors are regularized through a sigmoid function to obtain $\mathbf{i}_t$, $\mathbf{f}_t$, $\mathbf{o}_t \in \mathbb{R}^D$, respectively  representing the input gate vector (which weights the contribute of new information), the forget gate vector (which maintains old information) and the output gate. 
The $tanh$ function, instead, creates a vector on new candidate values, $\mathbf{\tilde{c}}_t\in \mathbb{R}^D$, that can be added to the state.

The i-th trajectory position $(x,y)^i_{t+1}$ is estimated considering the output state $\mathbf{h}^i_{t}$ and a bivariate Gaussian distribution at time $t$:
\begin{align}
	(\mu,& \sigma, \rho)^i_{t+1}= \tilde{W}\mathbf{h}^i_t,\\
	(x,y)^i_{t+1} &\sim \mathcal{N}(\mu^i_{t+1}, \sigma^i_{t+1}, \rho^i_{t+1}),
\end{align}
where $\mu^i_{t+1}$, $\sigma^i_{t+1}$ are the first and the second order moments of the Gaussian distribution, while $\rho^i_{t+1}$ represent the correlation coefficient. These parameters are obtained by a linear transformation of the output state  $\mathbf{h}^i_{t}$ with the matrix $\tilde{W}\in \mathbb{R}^{(5\times D)}$.

Given the i-th trajectory, the parameters are learned by minimizing the negative log-Likelihood loss:
\begin{align}
	L^i(W,W_x,\tilde{W})=-\sum_{t=1}^{T_{pred}} log\big(P\big((x,y)^i_t|(\mu,& \sigma, \rho)^i_t\big)\big)
	\label{eq:loss}
\end{align}

Even though these networks are a really powerful tool for modelling time-dependent phenomenon, they are not able to take into account for other factors that can influence the path of a person, such as interactions with other persons and interactions with static elements of the scene.

\paragraph{Encoding human-human interactions.}
A solution for modelling interactions between persons moving in the same space has been recently introduced in~\cite{alahi2016lstm}. Here, in order to model human-human interactions, at every time-step, the positions of all neighbors for the i-th trajectory are pooled through an occupancy grid of cells of size $m\times n$, $G[m,n]$. The occupancy matrix $O_t^i(m,n)$ is computed as follows:
\begin{align}
O_t^i(m,n)=\sum_{j\in G[m,n]} \mathbf{1}_{mn}[x_t^i-x_t^j, y_t^i-y_t^j],
\label{eq:olstm}
\end{align}
where $\mathbf{1}_{mn}$  is the indicator function, used to assign for each cell $[m,n]$ of the grid the corresponding $(x,y)^j_t$ trajectories. This matrix allows modelling the presence or absence of neighbors for each person.

This model is able to modify the trajectory in order to avoid immediate collision. However, in order to have a more smooth prediction a second model has been introduced, which simultaneously take into account for the hidden states of multiple LSTMs using a pooling layer, such as:
\begin{align}
H_t^i(m,n,:)=\sum_{j\in G_{m\times n}} \mathbf{1}_{mn}[x_t^i-x_t^j, y_t^i-y_t^j]\mathbf{h}_{t-1}^j.
\label{eq:slstm}
\end{align}

These two models do not consider the context in which a person is moving, and this may limit their application in real crowded spaces. For this reason, we introduce a context-aware pooling in which both human-human and human-space interactions are explicitly taken into account.

\begin{figure}
\centering
\includegraphics[width=0.8\textwidth]{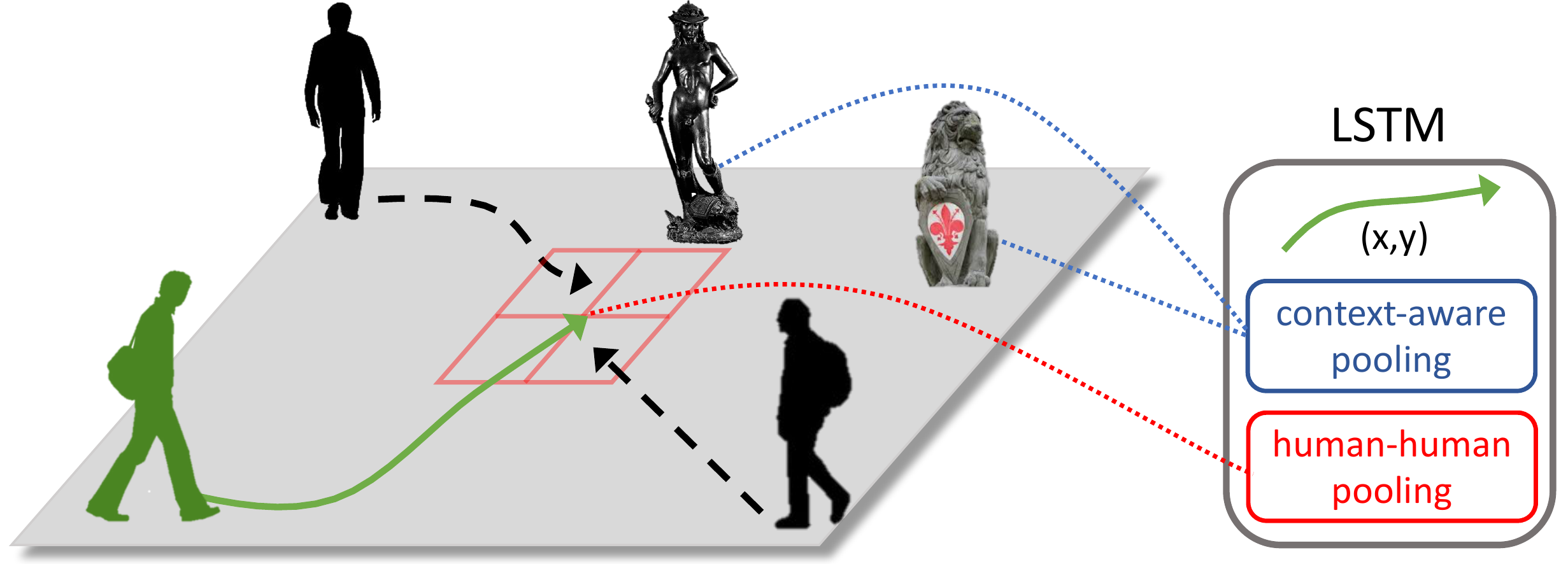}
\caption{Overview of our Context-aware Trajectory Prediction. The input of the LSTM are represented by the trajectory of the person under analysis (green), the grid of human-human interactions for the human-human pooling (red), and the context-aware pooling (blue). \label{fig:method}}
\end{figure}

\paragraph{Encoding human-human and human-space interactions.}
Considering the space in which a person is moving is fundamental to obtain a more accurate prediction. To this end, we first identify a set of static objects in the scene that can influence the human behavior. These points are manually defined and can comprise just some entry or exit points, but also more complex elements that can influence in a different way the behavior of a person (such as an artwork in a museum). An overview of our approach is given in Figure~\ref{fig:method}.

A naive approach for modelling also human-space interactions, can be obtained by modifying Eq.~\ref{eq:olstm} in order to include the static elements in the neighborhood of each person, as in the following:
\begin{align}
S_t^i(m,n,k)=\sum_{j\in G_{m\times n}} \mathbf{1}_{mn}[x_t^i-p_x^k, y_t^i-p_y^k],
\label{eq:oart}
\end{align}
where $(p_x,p_y)^k$ represents the coordinates of the k-th static object in the scene.
However, a drawback in this model is that each element in the space contributes equally to the prediction. 
Weighting equally the humans present in the neighborhood is useful to understand whether a trajectory is close to a collision or not. However, this same principle applied to the static objects in the scene is somewhat limited. Indeed, a human trajectory is influenced differently by humans or static objects.
For this reason we introduce a solution to model this latter case by defining for each person a vector containing the distances with respect to each element in the space:
\begin{align}
C_t^i(k)= \sqrt{(x_t^i-p_x^k)^2 +(y_t^j-p_y^k)^2}
\label{eq:ctx}
\end{align}
Differently from the naive solution of Eq.~\ref{eq:oart} in this way we can model how much each static elements can influence the path of a person. 

A straightforward way to include this kind of  interactions in our basic LSTM model can be obtained by simply modifying the input defined in Eq.~\ref{eq:xt} so as to include the representation defined in Eq.~\ref{eq:ctx}, such that:
\begin{align}
\mathbf{x}_t^i = \big[ \phi(X_t^i,W_x) ~~ \phi(C_t^i,W_C) \big]
\end{align}
Finally, depending on which approach we want to exploit for modelling human-human interactions, we can further extend the input to our model by concatenating the representations defined in  Eq.~\ref{eq:olstm} and Eq.~\ref{eq:slstm}, respectively:
\begin{align}
\mathbf{x}_t^i = \big[ \phi(X_t^i,W_x) ~~ \phi(C_t^i,W_C) ~~ \phi(O_t^i,W_O) \big]\\
\mathbf{x}_t^i = \big[ \phi(X_t^i,W_x) ~~ \phi(C_t^i,W_C) ~~ \phi(H_t^i,W_H) \big]
\end{align}
The loss function defined in Eq.~\ref{eq:loss} is also modified accordingly, introducing the set of parameters $W_O,W_H,W_C$ in the LSTM optimization.

\subsection{Trajectory Prediction}
At test time we consider a different set of trajectories, not observed during training time. In particular, we feed our model with the set of locations $\{x_t,y_t\}_1^{t_{obs}}$ of all the persons observed in the interval  $[1,t_{obs}]$. 
We then estimate the near future trajectories, $\{\hat{x}_t,\hat{y}_t\}_{t_{obs}+1}^{t_{pred}}$, of each person considering: 1) their path until time $t_{obs}$; 2) the path of other persons observed in the same period of time; and 3) the distances w.r.t. each static object in the space.

%-------------------------------------------------------------------------
\section{Experiments}\label{sec:experiments}
In this section, we report on a set of experiments to assess the performance of our Context-aware Trajectory Prediction.
We first describe the datasets used for the evaluation of our model and the settings used to train our model.
Then, we report a comparison between the state-of-the-art and different configurations of our model.

\subsection{Datasets and Evaluation Protocol}
\paragraph{UCY.} Our initial experiments have been conducted on the standard UCY~\cite{crowdbyex} dataset. 
This dataset contains three sequences, namely \emph{ZARA-01}, \emph{ZARA-02} and \emph{University}, which have been acquired in two different scenarios from a bird's eye view.
The first scenario, \emph{ZARA}, presents a moderately crowd condition. The \emph{University} scenario, on the other hand, is really crowd, with persons walking in different directions and group of persons, stuck in the scene.

Since this dataset does not provide any annotation regarding human-space interactions, we have annotated the two \emph{ZARA} sequences by manually identifying 11 points in the scene. These points are mainly localized in proximity of enter and exit locations.
%Moreover, in order to perform a more thorough analysis we also introduce a new challenging dataset, \emph{MuseumVisits}. 

\paragraph{MuseumVisits.} Our new dataset has been acquired in the hall of a big art museum using four cameras, with small or no overlap. The cameras are installed so as to observe the artworks present in the hall and capture groups of persons during their visits. Figure \ref{fig:pullfigure} shows three different views of the museum hall.
Differently from existing datasets, here we can observe rich human-human and human-space interactions. For example, persons walking together in a group and stopping in front of an artwork for a certain period of time.
Trajectories for $57$ distinct persons have been manually annotated along with some metadata, such as, the group one person belong to, the artwork a person is observing, etc. These metadata are not exploited in this work. 

Some statistics comparing the datasets used in our experiments are given in Table~\ref{tab:datasets}. The number of persons observed in the \emph{ZARA} sequences is higher with respect to our dataset, but the average length of their trajectories is really small. This is mainly because of the nature of this dataset, since persons enter and leave the scene continuously. Moreover, our dataset is slightly more crowd with an average number of persons  per frame of 17 instead of 14 and, most importantly, it includes richer interactions with static objects in the scene. 

\begin{table}
\caption{Statistics for the MuseumVisits  dataset and the ZARA sequences.  \label{tab:datasets}}
\centering
\begin{tabular}{|l|c|c|}
\hline
& \textbf{MuseumVisits} & \textbf{ZARA sequeces} \\\hline
Total number of persons & 57 & 204 \\\hline
Average trajectory length & 422 & 72 \\\hline
Minimum trajectory length & 26 & 7 \\\hline
Maximum trajectory length & 1019 & 617 \\\hline
Average number of person per frame & 17 & 14 \\\hline 
Interactions with scene elements & \cmark & \xmark \\\hline
\end{tabular}
\end{table}

\paragraph{Evaluation protocol.}
Experiments are conducted as follows: the trajectory of a person is observed for $3.2$ seconds, then a trained model is used to predict $4.8$ seconds. Trajectories are  sampled so as to retain one frame every ten; with a framerate of 0.4, this corresponds to observing 8 frames and predicting 12 frames, as in~\cite{Pellegrini2009,alahi2016lstm,suav2016eccv}. 
Results are reported in terms of \emph{Average displacement error}, which is calculated as the mean square error (MSE) in meters between the points predicted by the model $\{x_t^i,y_t^i\}$ and the ground-truth $\{g_t^i,g_t^i\}$:
\begin{align}
MSE= \sum_{i=1}^N\sum_{t=t_{obs}+1}^{t_{pred}} \frac{\sqrt{(x_t^i-g_t^i)^2 +(y_t^j-g_t^i)^2}}{N(t_{pred}-t_{obs})},
\end{align}
where $N$ is the total number of trajectories to be evaluated.
\subsection{Implementation Details}
In our Social-LSTM implementation~\cite{alahi2016lstm} we use the same configuration of parameters as in the original paper.
In particular, the embedding dimension for the spatial coordinates is set to 64 while the hidden state dimension is equal to 128, for all the LSTM models. 
The pooling size for the neighborhood of a person is set to $32$ and the dimension of the windows used for the pooling is $8\times8$. 
Each model has been trained considering a learning rate of 0.005, the RMSProp optimizer with a decay rate of 0.95, and a total of 1,600 epochs. All the solutions have been implemented using the TensorFlow framework and train/test have been performed on a single GPU. 
No further parameters are needed for our context-aware pooling since we consider the distance between a person and all the objects in the scene (this obviously depends only on the dataset under analysis). 

\begin{table}
\centering
\caption{Prediction errors on MuseumVisits and UCY (ZARA sequences). Results are reported in meters.\label{tab:res_sota}}
\scalebox{0.84}{
\begin{tabular}{|l|c|c|c|c|c||c|c|c||c|}
\hline
& \multicolumn{6}{c|}{\textbf{MuseumVisits}} & \multicolumn{3}{c|}{\textbf{UCY (ZARA sequences)}}\\
\hline
\textbf{Technique} & \textbf{Seq1} & \textbf{Seq2} & \textbf{Seq3} & \textbf{Seq4} & \textbf{Seq5} & \textbf{Avg} & \textbf{Seq1} & \textbf{Seq2} 
& \textbf{Avg} \\
\hline
\hline
LSTM   & \textbf{0.99} & 1.22 & 0.99 & 0.78 & 1.03 & 1.00 & 1.32 & 1.49 & 1.40 \\\hline
O-LSTM & 1.60 & 1.43 & 0.95 & 0.76 & 1.02 & 1.15 & 1.65 & 1.40 & 1.52\\\hline
S-LSTM & 1.68 & 1.26 & 0.94 & 0.75 & 0.88 & 1.10 & 1.30 & 1.37 & 1.34 \\\hline
\hline
Context-aware LSTM   & 1.36 & 1.14 & 1.21 & \textbf{0.49} & 0.82 & 1.00 & 1.21 & 1.37 & 1.29 \\\hline
Context-aware O-LSTM & 1.53 & \textbf{1.08} & \textbf{0.90} & 0.57 & \textbf{0.80} & \textbf{0.98} & \textbf{1.18} & 1.34 & 1.26\\\hline
Context-aware S-LSTM & 1.48 & 1.27 & 0.94 & 0.54 & 1.07 & 1.06 & 1.19 & \textbf{1.25} & \textbf{1.22}\\\hline
\end{tabular}
}
\end{table}

\begin{table}
\centering
\caption{Prediction errors on both MuseumVisits and UCY (ZARA sequences) using a different pooling for encoding human-space interactions. Results are reported in meters.\label{tab:res_other}}
\scalebox{0.81}{
\begin{tabular}{|l|c|c|c|c|c||c|c|c||c|}
\hline
& \multicolumn{6}{c|}{\textbf{MuseumVisits}} & \multicolumn{3}{c|}{\textbf{UCY (ZARA sequences)}}\\
\hline
\textbf{Technique} & \textbf{Seq1} & \textbf{Seq2} & \textbf{Seq3} & \textbf{Seq4} & \textbf{Seq5} & \textbf{Avg} & \textbf{Seq1} & \textbf{Seq2} & \textbf{Avg} \\
\hline
\hline
Context-aware LSTM (O)   & 1.66  & 1.55 & 1.26 & 0.81 & 1.20 & 1.30 & 1.30 & 1.40 & 1.35 \\\hline
Context-aware O-LSTM (O) & 1.93  & 1.44 & 1.32 & 0.71 & 1.15 & 1.31 & 1.37 & 1.40 & 1.39 \\\hline
Context-aware S-LSTM (O) & 1.66  & 1.55 & 1.26 & 0.58 & 1.20 & 1.25 & 1.31 & 1.29 & 1.30\\\hline
\end{tabular}
}
\end{table}

\subsection{Experimental Results}
In this section we discuss the results obtained with our solution compared to state-of-the-art. We re-implemented the three methods described in~\cite{alahi2016lstm}, namely LSTM, O-LSTM and Social-LSTM (S-LSTM). In particular, the solution based only on LSTM does not consider any kind of interactions while predicting the person trajectory. On the other hand, O-LSTM and S-LSTM are able to consider human-human interactions in their prediction. 
We then extend these three models with our context-aware pooling.
In our experiments we did not perform a fine-tuning using simulated trajectories, although according to~\cite{alahi2016lstm} this may help in reducing the overall prediction errors of each sequence of about 50\%. 

We divided the MuseumVisits dataset in five sequences and performed a five-fold cross validation while for the UCY datasets, we consider only the \emph{ZARA} sequences and performed a two-fold cross-validation. 
Table~\ref{tab:res_sota} shows experimental results for each sequence, as well as the average over all sequences.
For most of the MuseumVisits sequences, the combination of O-LSTM with our context-aware representation obtains the lowest errors. This can be motivated by the fact that persons in a museum usually moves from an artwork to another point of interest. 
Moreover, collisions are also quite limited in this kind of scenario since persons usually moves in group.
On the \emph{ZARA} sequences, instead, context-aware S-LSTM performs better, even though the combination with O-LSTM achieves the second-best result.
This is somehow expected, since in this scenario, a person usually walks alone from an entry point to an exit point and adjusts its path depending on the other persons. 

Finally, in Table~\ref{tab:res_other} we show some results obtained using a different implementation of our context-aware pooling which, similarly to O-LSTM, considers the human-space interactions by pooling the co-ordinates of neighbor static objects, as defined in Eq.~\ref{eq:oart}. The prediction errors are higher in this case. This confirm our assumption that each static object influences in different way the trajectory of a person, so weighting them equally, such as in the case of O-LSTM, is somehow limited.

\begin{figure}
\centering
\includegraphics[width=0.98\textwidth]{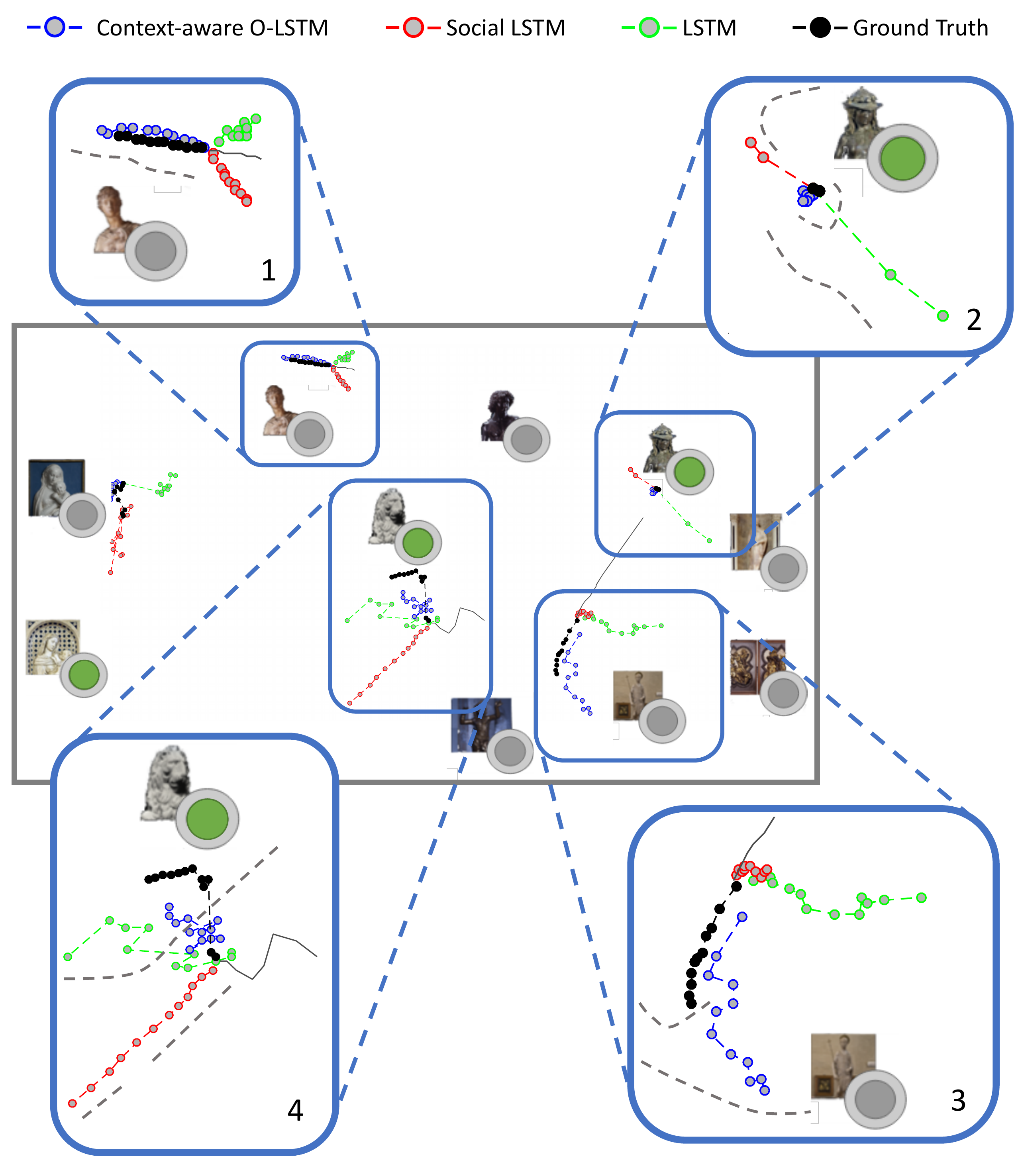}
\caption{Qualitative results on the MuseumVisits dataset, showing the trajectories predicted using LSTM (green circle), Social-LSTM (red circle), Context-aware O-LSTM (blue circle) and the Ground Truth (black circle). Four examples are highlighted in order to appreciate the differences between the three methods. Trajectories of other nearest persons are also shown (grey dashed lines). The green marker specifies that an artwork is observed by a person. \label{fig:qualitative}}
\end{figure}

\subsection{Qualitative Results}
Figure~\ref{fig:qualitative} shows some qualitative results of our model in comparison to prior work. 
We show four trajectories (in different colors), obtained using different methods. In black, we report the ground-truth trajectory. In green, a vanilla LSTM that does not use any information about the space and the other persons active in the scene. In red the output of Social-LSTM \cite{alahi2016lstm}, while our Context-aware model is shown in blue.

We highlight four examples where are depicted some static elements of the scene, such as a particular artwork, as well as the trajectories (in grey) of other subjects that can influence the path of our target.
In the first example, Context-aware O-LSTM is able to estimate  the real trajectory without being influenced by the artwork. Both LSTM and Social-LSTM are wrongly influenced by a close trajectory and fail in predicting the correct trajectory. More interesting is the second example which  shows the capability of our context-aware pooling to also exploit the static objects in the scene. On the contrary, solutions that do not model human-space interactions cannot stop the trajectory.
The third example is similar to the first one but our prediction is less accurate. This is mainly caused by the fact that our Context-aware LSTM wrongly estimated the motion model for this target.
The last example shows another interesting case in which the prediction is influenced by both static objects and other persons. In this case, our solution estimates a really noisy trajectory, close to the ground truth, while other methods are driven away.

%-------------------------------------------------------------------------
\section{Conclusion}\label{sec:conclusion}
We introduced a novel approach for trajectory prediction which is able to model interactions between persons in a scene as well as considering the rich semantic that characterize the space in which a person moves.
Inspired by the recent Social-LSTM model~\cite{alahi2016lstm}, which only considers human-human interactions, we introduce a more general formulation based on a ``context-aware'' pooling that is able to estimate how much the elements of the scene can influence the trajectory of a person.
We also introduce a new challenging dataset of trajectories, acquired in the hall of a big art museum, in a really crowded and rich scenario.

Experimental results on this new dataset and on a subset of the UCY dataset, demonstrate that considering both human-human and human-space interactions is fundamental for trajectory prediction. Our solution, indeed, obtains a lower prediction errors compared to other state of the art solutions. Additionally, we qualitatively show that our context-aware pooling is able to take into account for static object in the scene and stop a trajectory while also considering the influence of other trajectory in the neighborhood.

\bibliography{prediction}
\end{document}